\documentclass[letterpaper]{article} 
\usepackage{aaai25}  
\usepackage{times}  
\usepackage{helvet}  
\usepackage{courier}  
\usepackage[hyphens]{url}  
\usepackage{graphicx} 
\usepackage{natbib}  
\usepackage{caption} 
\usepackage{algorithm}
\usepackage{algorithmic}
\newcommand{\modelname}{ProIA}
\usepackage{booktabs} 
\usepackage{multirow}
\usepackage{amsmath}
\usepackage{color,xcolor}
\usepackage{amssymb}
\newtheorem{prop}{Proposition}
\usepackage{subfigure} 
\newtheorem{Lemma}{Lemma}
\newtheorem{Propapx}{Proposition}
\usepackage{newfloat}
\usepackage{listings}
\DeclareCaptionStyle{ruled}{labelfont=normalfont,labelsep=colon,strut=off} 
\floatstyle{ruled}
\newfloat{listing}{tb}{lst}{}
\floatname{listing}{Listing}
\title{Prompt-based Unifying Inference Attack on Graph Neural Networks}
\author{
    {Yuecen Wei\textsuperscript{\rm 1,\rm 2}, 
    Xingcheng Fu\textsuperscript{\rm 3}, 
    Lingyun Liu\textsuperscript{\rm 3}, 
    Qingyun Sun\textsuperscript{\rm 2}, 
    Hao Peng\textsuperscript{\rm 2}}, 
    Chunming Hu\textsuperscript{\rm 1,\rm 2}\thanks{Corresponding author}\\
}
\affiliations{
    \textsuperscript{\rm 1}School of Software, Beihang University, Beijing, China\\
    \textsuperscript{\rm 2}Beijing Advanced Innovation Center for Big Data and Brain Computing, Beihang University, Beijing, China\\
    \textsuperscript{\rm 3}Key Lab of Education Blockchain and Intelligent Technology, Ministry of Education, Guangxi Normal University, China\\  
    \{weiyc,sunqy,penghao,hucm\}@buaa.edu.cn,
    fuxc@gxnu.edu.cn,
    1295713045@stu.gxnu.edu.cn \\
}
\usepackage{bibentry}

\begin{document}

\maketitle

\begin{abstract}
Graph neural networks (GNNs) provide important prospective insights in applications such as social behavior analysis and financial risk analysis based on their powerful learning capabilities on graph data.
Nevertheless, GNNs' predictive performance relies on the quality of task-specific node labels, so it is common practice to improve the model's generalization ability in the downstream execution of decision-making tasks through pre-training. Graph prompting is a prudent choice but risky without taking measures to prevent data leakage.
In other words, in high-risk decision scenarios, prompt learning can infer private information by accessing model parameters trained on private data (publishing model parameters in pre-training, i.e., without directly leaking the raw data, is a tacitly accepted trend).
However, myriad graph inference attacks necessitate tailored module design and processing to enhance inference capabilities due to variations in supervision signals.
In this paper, we propose a novel \textbf{Pro}mpt-based unifying \textbf{I}nference \textbf{A}ttack framework on GNNs, named \textbf{\modelname}. 
Specifically, \modelname~retains the crucial topological information of the graph during pre-training, enhancing the background knowledge of the inference attack model.  
It then utilizes a unified prompt and introduces additional disentanglement factors in downstream attacks to adapt to task-relevant knowledge. 
Finally, extensive experiments show that~\modelname~enhances attack capabilities and demonstrates remarkable adaptability to various inference attacks.

\end{abstract}

\section{Introduction}
Real-world data benefits from the modeling of graph structures, which more effectively captures the inherent interconnected properties of the data, thereby enhancing the training process of models for Graph Neural Network (GNN) message passing~\cite{Kipf2017GCN, GAT, zhang2021GCN}.
In practical applications, besides conventional scenarios such as social recommendation~\cite{sharma2024survey} and traffic prediction~\cite{zhang2024survey}, GNNs have also demonstrated outstanding performance in high-risk scenarios such as fraud detection~\cite{innan2024financial} and disease prediction~\cite{boll2024graph}.
Despite this, a multitude of existing GNNs rely heavily on supervised learning, and their representational capacity is significantly hampered by the challenges associated with obtaining high-quality labels for specific domains~\cite{zhang2024bayesian,liu2024noisy,wu2022information}.
Self-supervised learning has consistently achieved superior performance across an increasing range of tasks, enhancing the generalization ability of GNNs.
Meanwhile, such graph learning models raise concerns about data security due to their ability to learn and extract knowledge. Many works in security have confirmed that pre-train models can lead to privacy leaks because of their inherent memory capabilities~\cite{duan2024membership}.
For instance, in early question-answering systems, personal information utilized for model pre-training could be acquired through straightforward prompt-based queries~\cite{carlini2021ExtracteData}. 
Furthermore, even in modern systems equipped with security and ethical modules, private information can still be extracted by implementing jailbreak techniques~\cite{li2023jailbreakingAttack}.
Existing prompt attacks against pre-train models have only been defined within natural language processing (NLP) models. In contrast, attacks on graph data with non-Euclidean structures have yet to be discussed.
\begin{figure}[t]
\centering
\includegraphics[width=0.45\textwidth]{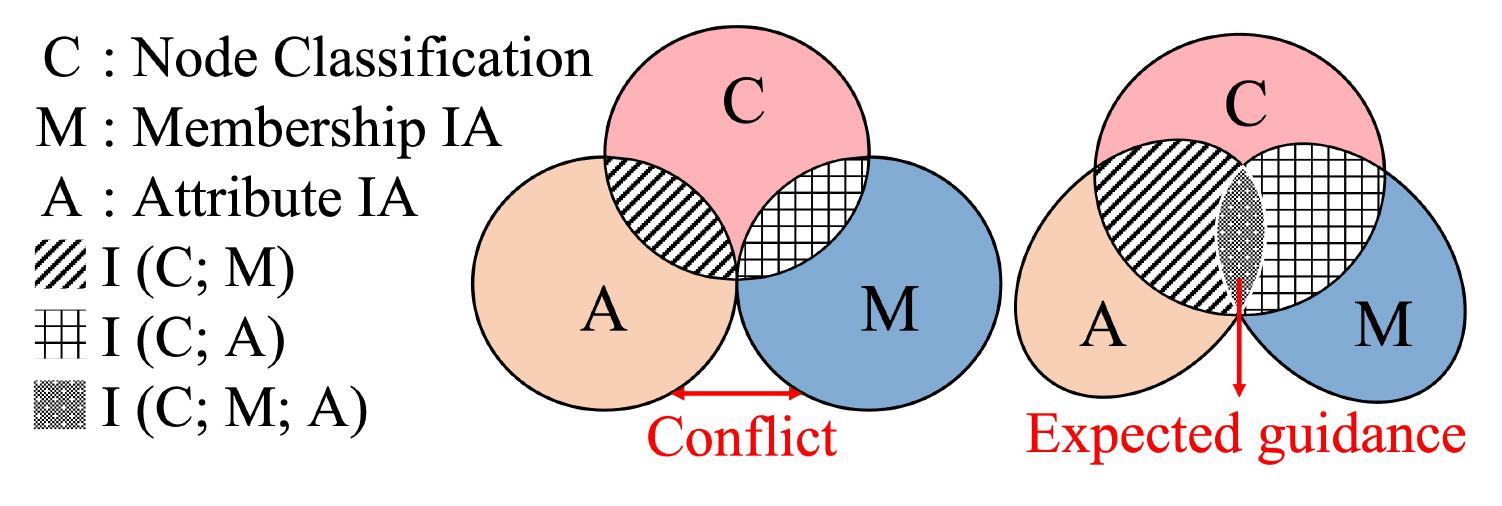} 
\caption{
The Venn diagram depicts the respective optimization targets for the tasks with the intersections. 
$I (C; M)$: increases training-test data disparity, 
$I (C; A)$: improving data fitting, and $I (C; M; A)$: weakening model generalization and increasing model sensitivity to specific attributes.}
\label{fig:motiv}
\end{figure}

In the evolution of models, state-of-the-art GNNs~\cite{zhiyao2024opengsl,liu2023multi,yuan2024environment} typically show positive, robust resistance and generalization capabilities when dealing with sparse and noisy data, thus effectively mitigating the issue of model overfitting. 
This results in privacy inference attacks that rely on model overfitting defects empirically failing to achieve satisfactory outcomes. 
However, even if the model avoids privacy leakage caused by overfitting, the adversary can still infer the target information by exploiting key and unique topological attributes of graph topology, such as scale-free and hierarchy~\cite{carlini2021ExtracteData, MIA2021,wei2024poincare}. 
Nevertheless, existing privacy inference attack methods are usually designed to guide a single attack independently.
The optimization targets for these classifications are different, especially when the adversary possesses multidimensional background knowledge, such as specific topologies. This makes it challenging to achieve satisfactory performance using uniformly guided attacks. 
As shown in Fig.~\ref{fig:motiv}, there is a conflict between simultaneously increasing the train-test distribution gap (achieving M) and improving data fitting capability (achieving A). In contrast, $I (C; M; A)$ illustrates the adversary’s expected goal. 

To investigate the capability of prompts in enhancing privacy inference attacks on graph data and to homogenize different attack tasks, it is non-trivial to address the following two challenges: $(1)$ 
The vulnerability of graph structures means that models can easily crash when faced with harmful inputs or noise, thus revealing the attack. 
And the extent to which prompts can enhance graph data attacks remains unknown. 
Therefore, \textbf{it is vital to determine what kind of prompt queries can reveal private data while remaining inconspicuous to the target model}. $(2)$ 
The most existing inference attacks necessitate the design of specific attack strategies. 
Integrating multifaceted information is challenging, as \textbf{extracting target information that matches downstream attack tasks to achieve adaptive inference attacks is difficult}.

\textbf{Present work.} 
To tackle these challenges, we propose a novel \textbf{Pro}mpt-based unifying \textbf{I}nference \textbf{A}ttack framework on GNNs, named \textbf{\modelname}, to enhance inference attacks and adaptive downstream attack tasks. 
Specifically, we design prompt queries for the target model to leak private information. Initially, we design information-theoretic principles during pre-training to guide the model in learning the essential structures of the graph, thereby enhancing the reachability of sensitive information. Subsequently, we construct prompt features through global-local contrastive learning to make the prompts appear as benign queries and make them unnoticeable to the target. 
Moreover, we design a disentanglement mechanism for the target model's output, allowing downstream tasks to infer the latent influence of corresponding variables from the enhanced posteriors provided by the prompt features, thereby adaptively achieving inference and further improving attack performance. 
Finally, extensive experiments validate the effectiveness of~\modelname.
Our contributions are summarized as:
\begin{itemize}
    \item We explore the potential for prompt-induced graph data leakage and subsequently utilize feature embedding in graph pre-training to incorporate more useful structural information and camouflage malicious prompt features. 
    \item Based on the prompt feature, we propose a novel \textbf{Pro}mpt-based unifying \textbf{I}nference \textbf{A}ttack framework on GNNs, named \textbf{\modelname}, 
    that incorporates a disentanglement strategy to extract latent factors and adaptively guide various downstream attack tasks.
    \item We conduct extensive experiments on five public datasets, and the results demonstrate that~\modelname~exhibits superior privacy inference capabilities and has a disruptive effect on several common defense mechanisms.
\end{itemize}

\section{Related Work}
\subsection{Graph Inference Attack}
Inference attacks have evolved from deep learning models~\cite{zarifzadeh2024low,  tramer2022truth, MIAoverfitting2017} to GNNs~\cite{TrustworthySurvey2022, zhang2023survey} by incorporating graph structures~\cite{yuan2024dynamic}. Due to the strong scalability of inference attacks in most applications, current research primarily focuses on transforming data formats (from Euclidean to non-Euclidean). 
\citet{duddu2020quantifying} and \citet{MIA2021} are the first to implement MIA on GNNs, incorporating graph structures into both target and shadow models, thereby establishing the pipeline for conducting MIA on graphs. 
\citet{conti2022label} utilized only label information to carry out MIA, reducing the adversary's need for background knowledge. 
Subsequently, 
\citet{AIA2022} and \citet{wu2021adapting} expanded node-level inference attacks to the graph level. 
\citet{wang2022group} analyzed the adversary's ability to infer attributes from a group perspective. 
\citet{olatunji2023does} proposed an iterative query strategy based on the feature propagation algorithm to enhance attribute inference attacks. 
Moreover, numerous defensive~\cite{dai2023unified, hu2022learning} efforts actively address adversaries with increasingly extensive background knowledge.

However, most existing works require the proposal of a particular framework to enhance a specific inference attack, with little attention given to unified guidance strategies for different inference attacks.

\subsection{Prompt Attack on Language Models}
Prompting~\cite{brown2020language, liu2023PPPSurvey} initially emerged in the field of NLP as directives designed for specific downstream tasks. It can serve as a substitute for fine-tuning to assist downstream tasks in retrieving relevant knowledge from pre-trained models. 
Privacy leakage during pre-training is typically associated with overfitting~\cite{overfitting2017, deng2024multilingual}. 
The erroneous association between overfitting and memorization has led many to believe that advanced models trained on large-scale data would not disclose information from their training data~\cite{carlini2021ExtracteData}. However, existing work has demonstrated the fallacy of this belief. 
They focus on enhancing the semantic information of structured prompt texts~\cite{liu2023autodan, yao2024fuzzllm, niu2024jailbreaking}, utilizing adversarial attacks to bypass Language Models (LMs) defense mechanisms~\cite{li2024drattack, shi2024optimization}, and designing gradient-enhanced strategies to generate efficient prompts~\cite{liu2024automatic}. 

However, the existing prompt designs for attacks focus on structured discrete data, making it challenging to adapt to the graph data structure. 
Moreover, graph prompts are an excellent alternative, yet they are not directly applicable to the pipeline of inference attacks. 

\section{Preliminary}

\subsection{Notations} 
We implement attribute inference attacks (AIA) and membership inference attacks (MIA) in GNNs with continuous and discrete attributes, respectively. 
Given a graph $G=\left ( V,E,\mathbf{X} \right )$ with node set $V=\left \{ v_1,\dots,v_N  \right \} $, edge set $E\in V\times V$ and node attribute matrix $\mathbf{X}  =\left \{ \mathbf{x}  _1,\dots,\mathbf{x}  _N  \right \} $, where $\mathbf{x}_i \in \mathbb{R}^d $ denote the feature vector of node $v_i \in V$. 
Let $\mathbf{A}_{ij} $ denotes the adjacency matrix and $\mathbf{A} _{ij}\in \left \{ 0,1 \right \} $ represents the connection between node $v_i$ and node $v_j$. 
For inference attack (IA), we focus on the node classification task, given the adversary's known labeled node set $V^{tr}$ and their labels $\mathcal{Y}^{tr}$. 
We aim to train a node classifier $\mathcal{F} _{\theta}$ that distinguishes in AIA and MIA to predict the sensitive labels $\mathcal{Y}^{te}$ of remaining unlabeled nodes $V^{te}=V \setminus V^{tr}$. 

\subsection{Inference Attack}
\noindent \textbf{Attribute Inference Attack. } 
The objective of attribute inference attacks is to train an attack model $\mathcal{F}_A$ by retraining the prediction results $\mathbf{h}_T$ of the target model $\mathcal{F}_T$ for specified queries $\mathbf{p} $, thereby inferring the sensitive attributes $\hat{\mathbf{x}} $ of nodes within the target training set $G_{T}^{tr}$. In real-world scenarios, such sensitive attributes might include social network users' gender, age, occupation, etc. 
Then, the attack model can be trained in the attack dataset $G_A$ by
\begin{equation}\label{eq:AIA}
\begin{aligned}
 \min _{\theta_A} \frac{1}{\left|G _A\right|} \sum_{G_i \in G_A} \mathcal{L}  \left(\mathcal{F} _A\left(\mathbf{x}_A^{tr}\right)_i^{G_i} , \mathbf{x} _i^{te}\right), 
\end{aligned}
\end{equation} 
where $\mathbf{x}_A^{tr}=\mathbf{h}_T^{\mathbf{p}}=\mathcal{F} _T(\mathbf{p}^{G_i})$ is the train-set embedding of $G_i$ from target model $\mathcal{F} _T$ in query $\mathbf{p}$, $\mathbf{x}^{te}$ is test set, $G_i$ is $k$-hop subgraph centered at node $v_i$, and 
$\mathcal{L} (\cdot )$ can be MSE loss or cross-entropy loss for different types of attributes. 

\noindent \textbf{Membership Inference Attack. } 
The objective is to infer whether a target individual was included in the training dataset of a target model $\mathcal{F}_A$, i.e., whether $\mathbf{x} \in G_T^{tr}$ or not, implying that MIA will lead to privacy leakage when $\mathcal{F}_A$ is trained on high-risk data. 

MIA consists of a target model $\mathcal{F}_T$, a shadow model $\mathcal{F}_S$, and an attack model $\mathcal{F}_A$, distinguishing it from AIA by an additional shadow model. Specifically, the function of $\mathcal{F}_A$ remains consistent with that in AIA, but the output of the target model $\mathbf{h}_T^{tr}$ does not directly guide $\mathcal{F}_A$. Instead, it is utilized to instruct $\mathcal{F}_S$ to learn to generate the attack dataset $G_A$ by mimicking the prediction behavior of $\mathcal{F}_T$, and its optimization objective is illustrated as: 
\begin{equation}\label{eq:MIA}
\begin{aligned}
\min _{\theta_S} \frac{1 }{\left|G_S^{tr}\right|} \sum_{G_i \in G_S^{tr}} \mathcal{L} \left(\mathcal{F} _S\left(G_i\right), \mathcal{F} _T\left(G_i\right)\right), 
\end{aligned}
\end{equation} 
where $\mathcal{F}(G_i)$ denotes the predicted label distribution of $G_i$.
Finally, the attack can be implemented using a Multilayer Perceptron (MLP). 
It is noteworthy that MIA is a binary classification task and typically relies on the assumption that $\mathcal{F}_T$ overfits to achieve the attack. However, most competitive models are designed to mitigate this issue effectively.

\noindent \textbf{Prompt Attack. } 
Prompt attacks macroscopically indicate that an adversary constructs specific inputs $\mathbf{p}$ to the target model $\mathcal{F}_T$, driving the model to search its memory $\theta_T $ for information related to the prompts and feed it back to the adversary $\mathcal{F}_A$. Specifically, 
the prompt attack~\cite{promptattack2023} in NLP aims to bypass the security mechanisms of language models, potentially causing them to reveal private information like phone numbers and addresses.
This paper then defines a similar concept called a ``prompt-based inference attack", which adheres to the aforementioned attack principles. 
It utilizes message passing in GNNs to establish unique connections between users. Even erased individual information can be extracted through prompting inference. 

\subsection{Graph Prompting} \label{sec:Grp-prom} 
Graph prompts are designed to rapidly adapt to downstream tasks by employing pre-training followed by prompts $\mathbf{p}$ for knowledge transfer. Specifically, to effectively extract task-specific prior knowledge from the pre-trained model, the node representation in the subgraph is generated using the ReadOut method as:
\begin{equation}\label{eq:prompt}
\begin{aligned}
\mathbf{h} _{query}=\textsc{ReadOut} \left \{ \mathbf{p}\odot \mathbf{h} \right \}, 
\end{aligned}
\end{equation}
where $\mathbf{h}$ is the node embeddings and $\odot $ denotes the element-wise multiplication. 
Additionally, node representations are fine-tuned during training to aid different downstream tasks. 
For different attacks, the core features of inference vary. 
For instance, AIA tends to focus on the expression of node attributes, whereas MIA emphasizes the overall data distribution.

\section{Prompt-based Inference Attack}
In this section, we will gradually introduce the detailed design of~\modelname, including the overall framework, the pre-training prompts, attack data generation, and the adaptive inference attack. 
See the appendix~\ref{sec:Alg} for~\modelname~'s algorithm.
\begin{figure*}[t]
\centering
\includegraphics[width=1\textwidth]{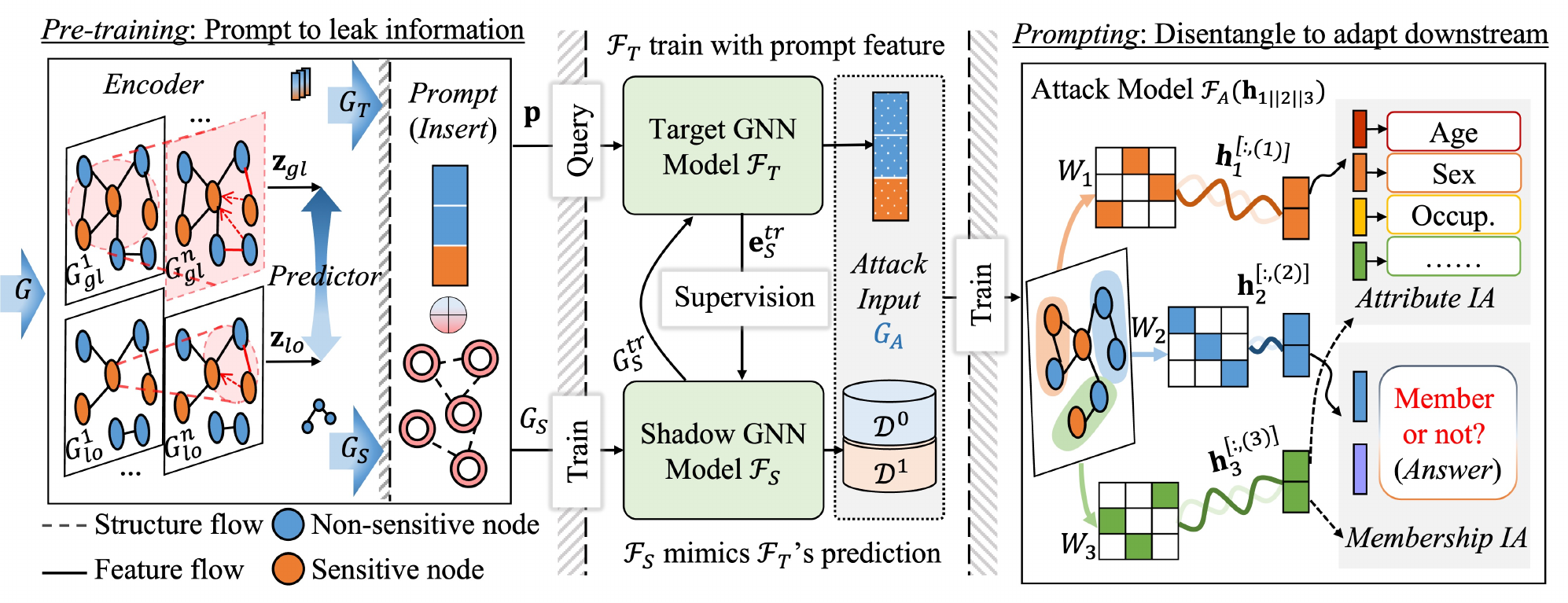} 
\caption{The overall framework of~\modelname.}
\label{fig:framework}
\end{figure*}

\subsection{Overview of~\modelname}
The overall framework of~\modelname~is illustrated in Fig.~\ref{fig:framework}. 
It consists of three components: pre-training, data generation, and prompt attack. 
$(1)$ Pre-training process, which aims to learn the critical topological structures and obscure malicious intents by contrasting local $G_{lo}$ and global $G_{gl}$ information, integrating rich information into the model to enhance its memory and attribute fitting capabilities. 
$(2)$ Attack data generation, where $\mathbf{p}$ obtained from the pre-trained model readout serves as prompt queries, forcing $\mathcal{F}_T$ to output information beneficial to the attack, which is then used as feature input for $\mathcal{F}_A$. For $\mathcal{F}_S$, it specifically refers to the module used in MIA to mimic $\mathcal{F}_T$'s inference and generate attack dataset supervision signals. 
$(3)$ Downstream inference attack, where an introduced disentangled representation module is used to further extract topology connections in $\mathbf{p}$ closely related to downstream tasks, enhancing information passed to guide adaptation to different attack tasks.

\subsection{Pre-training for leaking information}
\modelname~attempts to approach the conflict mentioned above from a unified perspective, avoiding direct bias towards either task.  
This aims to establish a more feasible influence for the propagation of sensitive attributes and enhance the robustness of the training data. 

\noindent \textbf{Sample generation. } 
We first quantify the similarity between nodes and their neighbors, which serves as a metric for initializing the construction of local subgraphs $G_{lo}$. Specifically, we set a threshold hyperparameter $t$ to control the connectivity of edges in the sampled local subgraphs, i.e., let $ \mathbf{s}  (\mathbf{x} _i, \mathbf{x} _j) $ denote the Jaccard similarity score between node $i$ and node $j$, if $\mathbf{s}  (\mathbf{x} _i, \mathbf{x} _j) \geq t$, then there exists an edge between $u$ and $v$ ($\mathbf{A} _{i,j}=1$). Otherwise, there is no edge ($\mathbf{A} _{i,j}=0$). 
Subsequently, the subgraph is augmented by adding edges through random Bernoulli sampling. Therefore, the positive sample is denoted as $\mathbf{x}$, and the negative sample $\hat{\mathbf{x } } $ is generated by randomly permuting the node features of $G_{gl}$. In the joint learning process with the contrastive samples, $G_{lo}$ enhances the model training's robustness against perturbations through reweighting. 

\noindent \textbf{Information bottleneck constraints.} 
As mentioned above, a pre-trained model that captures the key structures can provide diversity for information extraction. 
Therefore, we utilize information-theoretic techniques to guide the aggregation of critical information, i.e., the information bottleneck (IB) principle~\cite{wu2020graph}.

Let a local subgraph be composed of node $v_i$ and its $k$-hop neighbors containing the relevant data, assuming independence from the rest of the graph. 
Consequently, the IB constraint can approximate the globally optimal representation space within an iteratively refined Markov chain. 
Furthermore, to approximate the solution of $I(C,M,A)$ in Fig.~\ref{fig:motiv}, we need to optimize by: 
\begin{equation}
\begin{aligned}
&\min_{\theta } -I(\mathbf{Z};\mathbf{Y})+\beta_{A,M} I(G,\mathbf{Z}) \\
 =&\min_{\theta } -I(\mathbf{Z}_\mathbf{X};\mathbf{Y}^C)+\beta_{A,M} I(G;\mathbf{Z}_{At},\mathbf{Z}_{Me}),
\end{aligned}
\end{equation} 
where $\left \{ \mathbf{Z}_{At},\mathbf{Z}_{Me} \right \} \in \mathbf{Z}_\mathbf{X}$ denote the representations that facilitate attribute fitting and member information propagation, $I(\cdot )$ represents mutual information, $\beta$ is the Lagrangian parameter to balance the terms, and $\mathbf{Y}^C$ is the label of node classification. 
According to those above, after instantiating the IB in the Markov chain, we only need to optimize distributions $\mathbb{P} (\mathbf{Z}_{At}^{l}| \mathbf{Z}_{At}^{l-1},\mathbf{Z}_{Me}^{l})$ and $\mathbb{P} (\mathbf{Z}_{Me}^{l}| \mathbf{Z}_{At}^{l-1},\mathbf{A}), l\in L$  to capture the local dependencies between nodes for easier facilitating parameterization and optimization. 

To achieve a more precise attainment of the final optimization objective through local optimization, we introduce variational boundaries to optimize the model~\cite{alemi2016deep}. 
The lower bound of $I(Z_X;Y^C)$ and the upper bound of $I(G;\mathbf{Z}_{At},\mathbf{Z}_{Me})$ are inspired by~\citet{wu2020graph}.

\begin{prop} \label{pr:prop1}(The lower bound of $I(Z_X;Y^C)$).
For any variational distributions $ \mathbb{Q}_{1}\left(\mathbf{Y} ^C \mid \mathbf{Z} _{\mathbf{X} }^{L}\right)$ and $\mathbb{Q}_{2}(\mathbf{Y} ^C)$: 
\begin{equation}\label{eq:low_bound}
\begin{aligned}
I\left(\mathbf{Y} ^C ; \mathbf{Z} _{\mathbf{X} }^{L}\right) \geq 
1
+\mathbb{E}\left[\log \frac{ \mathbb{Q}_{1}\left(\mathbf{Y} ^C \mid \mathbf{Z} _{\mathbf{X} }^{L}\right)}{\mathbb{Q}_{2}(\mathbf{Y} ^C)}\right]& \\
-\mathbb{E}_{\mathbb{P}(\mathbf{Y} ^C) \mathbb{P}\left(\mathbf{Z} _{\mathbf{X} }^{L}\right)}\left[\frac{  \mathbb{Q}_{1}\left(\mathbf{Y} ^C \mid \mathbf{Z} _{\mathbf{X}}^{L}\right)}{\mathbb{Q}_{2}(\mathbf{Y} ^C)}\right]&
\end{aligned}
\end{equation} 
\end{prop} 

\begin{prop}\label{pr:prop2} 
(The upper bound of $I(G;\mathbf{Z}_{At},\mathbf{Z}_{Me})$).
Two randomly selected indices $\mathcal{I}_ {a}, \mathcal{I}_ {m} \subset L$, are independent ($G \perp \mathbf{Z}_{At}^{L} \mid \left\{\mathbf{Z}_{At}^{l_a}\right\} \cup\left\{\mathbf{Z}_{Me}^{l_m}\right\}$) based on the aforementioned Markov chain, where $l_a \in \mathcal{I} _a$, $l_m \in \mathcal{I} _m$. 
Then for any variational distributions $\mathbb{Q}\left(\mathbf{Z}_{At}^{l}\right)$ and $\mathbb{Q}\left(\mathbf{Z}_{Me}^{l}\right)$:

\begin{equation}\label{eq:up_bound}
\begin{aligned}
I(G;\mathbf{Z}_{At},\mathbf{Z}_{Me})
&\leq I \left(G;\left\{\mathbf{Z}_{At}^{l_a}\right\} \cup\left\{\mathbf{Z}_{Me}^{l_m}\right\}\right) \\
&\leq \sum_{l \in \mathcal{I}_a} \mathcal{T} _{At}^{l}+\sum_{l \in \mathcal{I}_m} \mathcal{T} _{Me}^{l} ,
\end{aligned}
\end{equation} 
\begin{align}
\text {where } 
&
\mathcal{T} ^{l}_{At}=\mathbb{E}\left[\log \frac{\mathbb{P}\left(\mathbf{Z}_{At}^{l} \mid \mathbf{Z}_{At}^{l-1}, \mathbf{Z}_{Me}^{l}\right)}{\mathbb{Q}\left(\mathbf{Z}_{At}^{l}\right)}\right]
, \\
&
\mathcal{T} ^{l}_{Me}=\mathbb{E}\left[\log \frac{\mathbb{P}\left(\mathbf{Z} _{Me}^{l} \mid \mathbf{A}, \mathbf{Z} _{At}^{l-1}\right)}{\mathbb{Q}\left(\mathbf{Z}_{Me}^{l}\right)}\right]
.
\end{align}

\end{prop} 
The proofs are provided in the appendix~\ref{sec:proof}.

To estimate $\mathcal{T} _{At}^l$, let $\mathbb{Q}\left(\mathbf{Z}_{At}^{l}\right)$ be a Gaussians distribution $q(\mathbf{Z}_{At})$, $p(\mathbf{Z}_{At},\mathbf{Z}_{Me})$ uses the sampled by $\mathbf{Z}_{At}$:
\begin{equation}
\mathcal{L}_{A}=\sum_{v\in V} \mathrm{KL} \left[\left(p(\mathbf{Z}_{{At},v},\mathbf{Z}_{{Me},v}|\mathbf{Z}_\mathbf{x}) \| p(\mathbf{Z}_{{At},v})\right)\right].
\end{equation} 
Similarly, the estimation of $\mathcal{T} _{Me}^l$:
\begin{equation}
\mathcal{L}_{M}=\sum_{v\in V} \mathrm{KL} \left[\left(p(\mathbf{Z}_{{Me},v},\mathbf{Z}_{{At},v}|\mathbf{Z}_\mathbf{x}) \| p(\mathbf{Z}_{{Me},v})\right)\right].
\end{equation} 
Then, the objective function of IB is:
\begin{equation}
\mathcal{L}_{IB}\!=\!-\!\sum_{v \in V} \!\mathrm{CE}\!\left(f(\mathbf{Z} _{\mathbf{X},v }^{L}) ; \mathbf{Y} ^C\right)\!+\!\beta_A \mathcal{L}_A\!+\!\beta_M \mathcal{L}_M,
\end{equation} 
where $f(\cdot )$ is a classifier, $\mathrm{CE} $ represents the cross-entropy loss about the upper bound. 

\noindent \textbf{Contrastive training.} 
We employ a graph convolutional network as the encoder, resulting in the node representation $\mathbf{z}$ with IB.
For the representations $G_{lo}$ and $G_{gl}$ denoted as $\mathbf{z}_{lo}$ and $\mathbf{z}_{gl}$, to capture the intricate interactions between the features of the local and global representation spaces, we obtain a unified representation via a Bilinear transformation layer: 
\begin{align} \label{eq:bilinear}
&\mathcal{P} =  \sigma (\mathbf{z} _{lo1}^{T})\mathbf{W} \mathbf{z}_{gl}+\mathbf{b}, \\ 
&\mathcal{\hat{P} } =  \sigma (\mathbf{z} _{lo2}^{T})\mathbf{W} \mathbf{\hat{z} }_{gl}+\mathbf{b},
\end{align}
where $\mathbf{W}$ is a learned weight matrix and $\mathbf{b}$ represents a learned bias vector. $\sigma (\cdot )$ is a nonlinear activation function. 
The model parameter $\phi$ is trained to maximize the mutual information between the global and local representations. Consequently, the objective function of the binary cross-entropy loss is expressed as: 
\begin{equation} \label{eq:losscl}
\begin{aligned}
 \mathcal{L} _{CL}=- \frac{1}{N} \sum_{i=1}^{N} \left[\log(\mathcal{P}) +  \log(1 - \mathcal{\hat{P} }) \right].
\end{aligned}
\end{equation}
Finally, the loss function for the entire pre-training is: 
\begin{equation} \label{eq:loss1}
\begin{aligned}
\mathcal{L} _1=\alpha \mathcal{L}_{CL}+(1-\alpha )\mathcal{L}_{IB},
\end{aligned}
\end{equation}
where $\alpha$ is a trade-off hyperparameter.

\subsection{Attack Data Generation}
We aim to design a unified prompt feature to enhance attack capabilities.
The attack processes for both attribute inference and membership inference follow established methodologies from previous research, and the manifestation forms of the prompt features are shown below. 

\noindent \textbf{Attribute Inference.} 
The training set $G_{A}^{tr}$ for $\mathcal{F}_A$ is derived from the adversary's background knowledge, while the test set $G_{A}^{te}$ is obtained from the output of $\mathcal{F}_T$. Specifically, given the known parameters $\phi$ of the pre-trained model, the adversary designs prompt features $\textbf{p}$ for $\mathcal{F}_T$ based on Eq.~\ref{eq:prompt} and initiates queries to the model. The output posteriors $\mathbf{h}$ are then used as input for $\mathcal{F}_A$. 

\noindent \textbf{Membership Inference.} 
In addition to $\mathcal{F}_T$, extra shadow models $\mathcal{F}_S$ in MIA are utilized to construct the training set $G_{A}^{tr}$ used for training $\mathcal{F}_A$. Specifically, the adversary leverages background data $G_S(\mathbf{p})$ to train a model $\mathcal{F}_S$ that has a similar architecture to $\mathcal{F}_T$. The node representations are obtained from $\mathcal{F}_S$, with $G_{S}^{tr}$ labeled as $1$ and $G_{S}^{te}$ labeled as $0$, thereby forming $G_{A}^{tr}=\mathbf{h}$. Furthermore, $G_{A}^{te}=G_T$.

\subsection{Prompting for adapting Inference Attack with disentanglement}
From the previous phase, AIA obtains a vector matrix about the node representation optimized on the attack model, and MIA gets an extra set of datasets to train the attack model. 
Then~\modelname~proposes a task-specific learnable prompt method for more effective knowledge transfer. 

\textbf{Prompt disentanglement.}
To ensure that $\mathcal{F}_A$ obtains guidance information that is more directly relevant to downstream tasks during inference,~\modelname~integrate a disentanglement mechanism into $\mathcal{F}_A$ to isolate key factors affecting the structure in prompt features. 
Specifically, in AIA, nodes are influenced by various attributes and establish potential connections. We aim to set up $k$ fixed virtual channels for the inter-attribute connections and learn the disentangled representation of latent factors. 
First, prompt features are pre-corrected for offsets through transformation and then mapped into channels, forming neighbors based on single/multiple attributes: 
\begin{equation}\label{eq:disenmapping}
\begin{aligned}
\mathbf{z}_{i,k}=\frac{ \sigma (\mathbf{W}_{k}^T (\mathrm{MLP} (\mathbf{p}_i) \odot\mathbf{h}_i)  + \mathbf{b} _k)}{ \left \| \sigma (\mathbf{W}_{k}^T (\mathrm{MLP} (\mathbf{p}_i) \odot \mathbf{h}_i)  + \mathbf{b} _k)  \right \|_2  } ,
\end{aligned}
\end{equation}
where $\mathbf{W}_{k}$ and $\mathbf{b} _k$ are the parameters of channel $k$. 
After the initialization above, the similarity weights between nodes $i$ and their neighbors $j$ are calculated within each channel, forming the edge probability between nodes and reweighting the nodes: 
\begin{equation} \label{eq:disen_reweigt}
\begin{aligned}
\hat{\mathbf{z}} _{j,k} = \mathbf{z}_{j,k}\odot \sum_{j:(i,j)\in G_A}\textsc{Softmax}(\mathbf{z} _{j,k}^T \mathbf{z}_{i,k}^{(t-1)})/\tau ,
\end{aligned}
\end{equation}
where $\tau$ is a hyperparameter that regulates the tightness of the cluster centers. 
Subsequently, the latent factor representations within the channel are updated to the nodes, resulting in disentangled representations: 
\begin{equation}\label{eq:disen_repres}
\begin{aligned}
\mathbf{d} _k^t=\frac{\mathbf{z} _{i,k} + \hat{\mathbf{z}} _{j,k}}{\left \| \mathbf{z} _{i,k} + \hat{\mathbf{z}} _{j,k} \right \| _2}, 
\end{aligned}
\end{equation}
where $t$ is a hyperparameter for the number of iterations in  Eq.~\ref{eq:disen_reweigt} and~\ref{eq:disen_repres}, aiming to search for the largest cluster in each subspace iteratively. Each neighbor is approximated to belong to only one subspace cluster. 

\noindent \textbf{Attack implementation.}
After multiple iterations, a multi-level disentangled representation $\textbf{d}$ is formed. This representation, once normalized, is used as prior guidance for training $\mathcal{F}_A$ in the MLP. The objective function is as follows: 
\begin{equation}\label{eq:loss2}
\begin{aligned}
\mathcal{L}_2 = \frac{1}{N} \sum_{i=1}^{N} \left( -\text{log} {p_i^A} -\text{log} {q_i^d}\right)+\lambda \mathcal{D} _\mathrm{KL}(\mathbb{P} ^A||\mathbb{Q} ^d),
\end{aligned}
\end{equation}
where $p_i^A=\mathcal{F}_A (\mathbf{\tilde{d} } _i\mathbf{h}_i )$ and $q_i^d=\textsc{Softmax}(\mathbf{W}^T \mathbf{d}_i+b)$, their KL divergence is employed further to constrain the impact of disentanglement factors on attack inference. 

\begin{table*}[ht]
\centering
\begin{tabular}{ccccccccc|ccccc}
\toprule
&
\multirow{2}{*}{\raisebox{-1ex}{Model}}
& \multicolumn{2}{c}{Cora} 
& \multicolumn{2}{c}{Facebook} 
& \multicolumn{2}{c}{Lastfm}
& \multicolumn{1}{c}{Imp.}
& \multicolumn{2}{c}{Bail}
& \multicolumn{2}{c}{Pokec-n}
& \multicolumn{1}{c}{Imp.}
\\ \cmidrule{3-8} \cmidrule{10-13}
&
&  ACC &  F1 &  ACC &  F1 &  ACC &  F1 &Avg. &  ACC &  F1 &  ACC &  F1 &Avg. 
\\ \midrule
\multirow{3}{*}{\rotatebox{90}{Vanilla}} &
GCN   &82.03   &81.60   &59.02   &58.91   &61.84   &60.40 &   &51.81   &51.80   &70.76   &59.55 &  \\
&GAT    &85.88   &85.70   &60.72   &55.17   &64.84   &64.11  &-  &50.61   &34.01   &71.24   &59.27  &-   \\
&SAGE    &73.77   &73.94   &60.01   &45.22   &61.51   &51.73  &  &52.96   &52.40   &70.86   &60.10  &   \\
\midrule
\multirow{3}{*}{\rotatebox{90}{Pre-tr.}} &
GCN    &56.14  &56.17  &55.22  &54.25  &42.55  &41.72 &-16.29   &\underline{55.14}  &\textbf{53.61}  &70.71  &58.57   &$\uparrow$1.4  \\
&STABLE    &63.10  &57.45  &61.58  &61.44  &67.13  &64.56  &-4.76   &54.29  &53.39  &NA  &NA   & -4.63 \\
&\textbf{\modelname}$_{\mathbf{p}}$    &89.47  &89.54  &\underline{62.88}  &\underline{62.81}  &70.40  &68.63  &$\uparrow$6.66    &54.41  &53.31  &70.51  &59.02   &$\uparrow$ 0.85   \\
\midrule
\multirow{3}{*}{\rotatebox{90}{Prompt}} &
\textbf{\modelname}$_{\mathbf{d}}$    &81.42  &81.09  &61.42  &61.34  &65.91  &64.77  &$\uparrow$2.03  &52.84	 &51.79   &70.69  &58.56   &$\uparrow$0.01   \\
&GCN    &80.95  &80.68  &53.80  &51.94  &43.15  &43.28   & -8.33   &53.67  &52.23  &72.93  &68.75    &$\uparrow$3.43  \\
&STABLE    &62.80  &56.56  &61.09  &60.92  &66.56  &63.32   &-5.43    &54.10  &\underline{53.43}  &NA  &NA    &-4.70  \\ 
\midrule
\multirow{3}{*}{\rotatebox{90}{\textbf{\modelname}}} 
&\textbf{GCN}    &90.18  &90.01  &\textbf{62.89}  &\textbf{62.85}  &\underline{70.45}  &\underline{68.61}   &   &\textbf{55.19}  &52.81  &74.85  &70.48   &   \\
&\textbf{GAT}    &\textbf{97.62}  &\textbf{97.60}  &60.99  &60.72  &66.09  &59.83     &$\uparrow$7.74  &52.03   &44.37    &\textbf{78.44}  &\textbf{77.06}   &$\uparrow$5.60   \\
&\textbf{SAGE}    &\underline{90.62}  &\underline{90.63}  &53.48  &51.79  &\textbf{80.05}  &\textbf{71.28}  &  &52.60   &48.35     &\underline{74.92}  &\underline{71.36}   &   \\
\bottomrule
\end{tabular}
\caption{Summary results of accuracy, Weighted-F1 and average improvement performance. (NA represents memory limitation)}
\label{tab:summaryResult}
\end{table*}

\begin{figure*}[ht] 
    \begin{minipage}{0.32\textwidth}
        \centering      
        \includegraphics[width=1\textwidth]{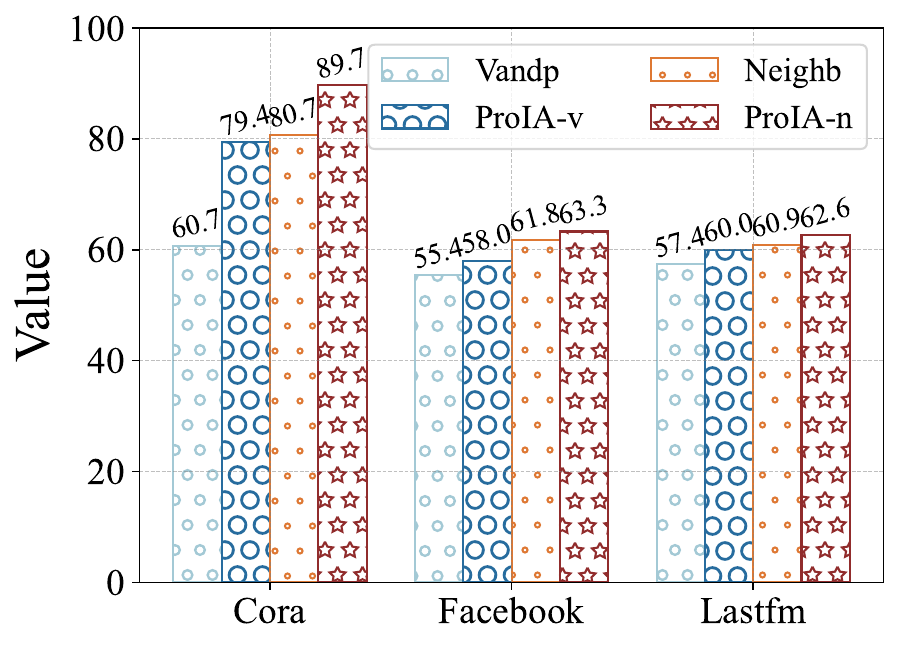}
    \end{minipage}
    \begin{minipage}{0.32\textwidth}
        \centering
        \includegraphics[width=1\textwidth]{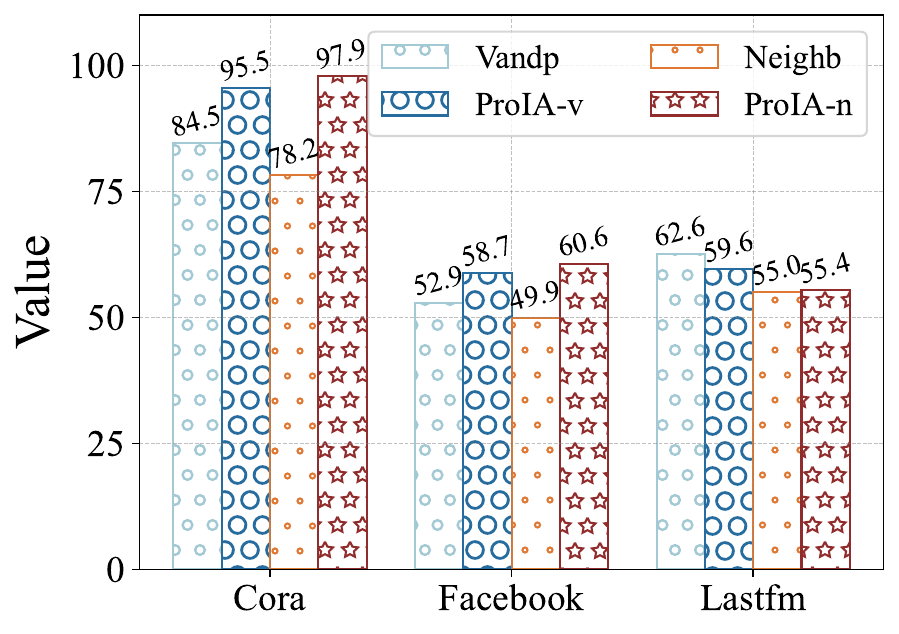}
    \end{minipage}
    \begin{minipage}{0.32\textwidth}
        \centering
        \includegraphics[width=1\textwidth]{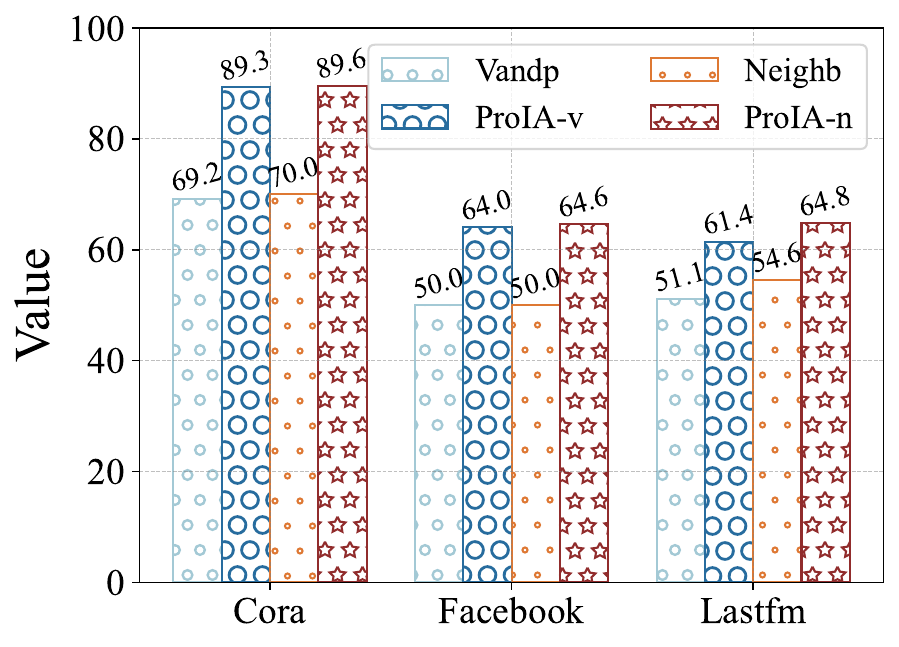}
    \end{minipage}

\centering
\caption{Attack AUC-ROC scores of GCN, GAT, and SAGE (from left to right) against defended models for MIA.}
\label{fig:defend_MIA}
\end{figure*}

\section{Experiment}
In this section, we conduct extensive experiments on five real-world datasets to validate the attacking capability of~\modelname\footnote{Code available: https://github.com/RingBDStack/ProIA}. 
We introduce the experimental setup, present the results, and provide a detailed analysis. 

\subsection{Experimental settings}
\textbf{Datasets.} 
\textit{For the MIA}, we employ three widely-used graph datasets. 
\textbf{Cora}~\cite{Kipf2017GCN} is a citation network of academic papers.
\textbf{Facebook}~\cite{leskovec2012learning} describes the social network of relationships between social media pages.
\textbf{Lastfm}~\cite{rozemberczki2020characteristic} is a social network reflecting users' musical interests. 
\textit{For the AIA}, we utilize two datasets that are labeled with sensitive attributes.
\textbf{Bail}~\cite{agarwal2021towards} is a U.S. bail dataset. 
\textbf{Pokec-n}~\cite{takac2012data} is a social network extracted from the Slovakian social network Pokec. 
See the appendix~\ref{sec:Dataset} for more details. 

\noindent \textbf{Baselines.}
We established three backbone models (GCN, GAT, and SAGE) to test the expressiveness of~\modelname~in different scenarios. 
\textbf{STABLE}~\cite{li2022reliable}, an additional model, shows~\modelname's utility in extracting hints from robust models using a contrastive learning framework.
Furthermore, we validated~\modelname's capability in defense models (\textbf{Vandp}~\cite{MIA2021}, \textbf{Neigbh}~\cite{MIA2021}, \textbf{PPGL}~\cite{hu2022learning}). 
See the appendix~\ref{sec:baseline} for more details.

Based on the backbones and attack backgrounds, the methods are divided into four categories: 
(1) \textbf{Vanilla} represents models used by both $\mathcal{F}_T$ and $\mathcal{F}_S$, with $\mathcal{F}_A$ being an MLP.
(2) \textbf{Pre-tr.} refers to obtaining hint features through pre-training on different models, applied to the training of $\mathcal{F}_T$ or $\mathcal{F}_S$ with GCN as the backbone, using an MLP as $\mathcal{F}_A$.
(3) \textbf{Prompt} introduces a disengagement mechanism for $\mathcal{F}_A$ in pre-training.
(4) \textbf{\modelname} demonstrates the proposed method's attack capability on targets in different scenarios. 

\noindent \textbf{Settings.}
To maximize its advantages, we set the number of layers of the baseline to $2$ and the disentangled mechanism layers in~\modelname~to $5$. 
The attack model's learning rate and iteration number are set to $0.01$ and $100$, and the remaining modules and methods are set to $1e-4$ and $200$. 
Common parameters include a representation dimension of $256$ and the Adam optimizer. 
The privacy-preserving model employs a unified privacy budget of $0.2$. 
All other settings use default optimal values. 
See the appendix~\ref{sec:setting} for more details.

\subsection{Performance Evaluation}\label{sec:performance}
Based on the categorization above, we uniformly conducted comprehensive performance verification and ablation experiments for~\modelname.
We then compared the inference capability of~\modelname~against commonly used privacy protection models. Finally, we design the case study to demonstrate the information extraction ability of~\modelname~during the pre-training and the downstream attack.

\begin{figure}[t]
{
\subfigure[Cora on GAT.]{
    \begin{minipage}{0.22\textwidth}
        \centering      
        \includegraphics[width=1\columnwidth]{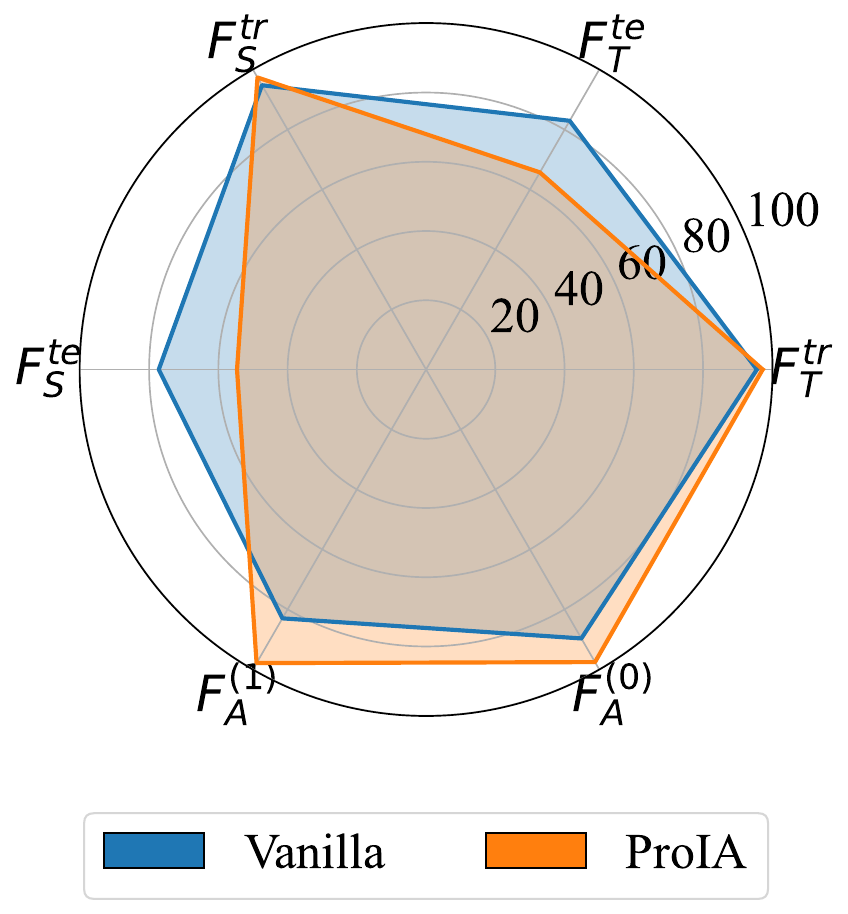}
        \label{fig:ca}
    \end{minipage}
}
\subfigure[Cora on Sage.]{
    \begin{minipage}{0.22\textwidth}
        \centering
        \includegraphics[width=1\columnwidth]{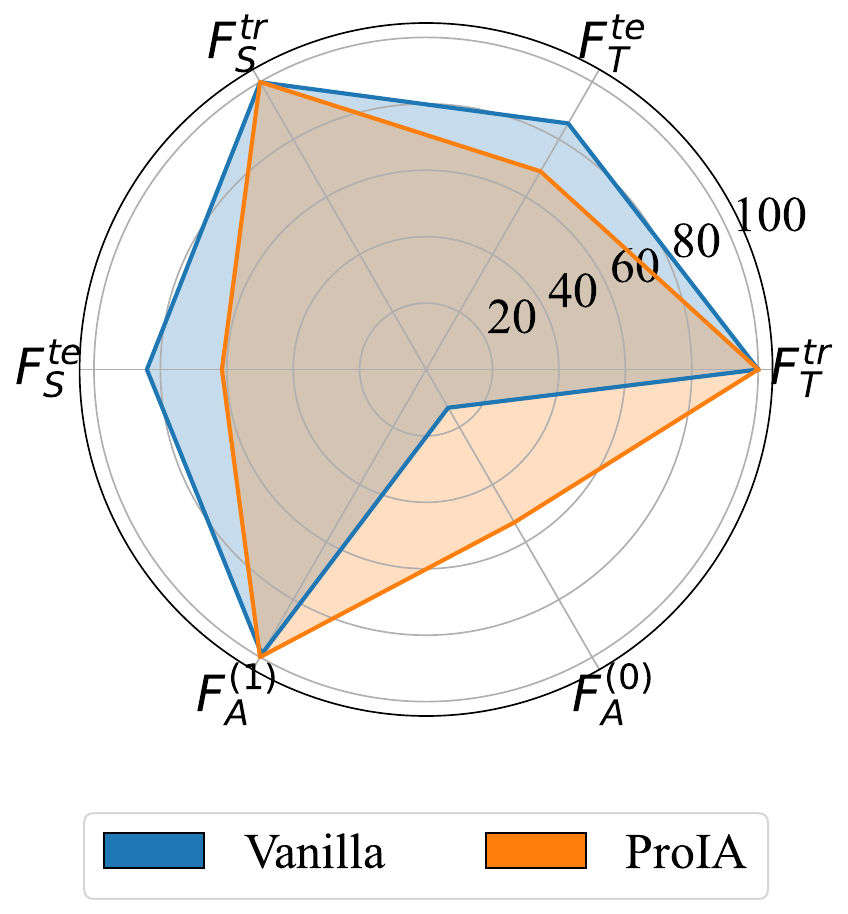}
        \label{fig:cb}
    \end{minipage}
}

\centering
\caption{Case study.}
\label{fig:casestudy}
}
\end{figure}

\noindent \textbf{Overall performance and ablation study.}
As shown in Tab.~\ref{tab:summaryResult}, we employed multiple metrics and devised various method variants. The average increase in~\modelname's attack accuracy indicates superior performance in both MIA and AIA. 
Specifically, compared to Vanilla,~\modelname~achieved a maximum accuracy improvement of 18.54\% for Lastfm in SAGE. Although~\modelname~is slightly disadvantaged under a few backbones, it maintains a substantial advantage by flexibly adapting the attack scenarios.

For the ablation studies involving Pre-tr and Prompt, our method~\modelname$_\mathbf{p}$ represents removing the disentanglement mechanism, and~\modelname${_\mathbf{d}}$ indicates that prompt features are not used as background knowledge, as shown in Tab.~\ref{tab:summaryResult} and Fig.~\ref{fig:ablationstudyAIA}. 
Both approaches show varying degrees of attack improvement compared to Vanilla, with single modules sometimes achieving sub-optimal results. 
This demonstrates their effectiveness and that not all prompt features used as queries can assist in the attack. 
Furthermore, data issues like label bias and noise within the dataset may understandably lead to varying module performance.
Overall,~\modelname~successfully extracts critical information for inference attacks and provides adaptive guidance for downstream attacks.

\noindent \textbf{Attack performance in defended models.}
To ensure fairness in model settings, we evaluated the impact of inserting defense methods into Vanilla and~\modelname~on attacks against the protected target model, with the results illustrated in Fig.~\ref{fig:defend_MIA} and~\ref{fig:defend_AIA}. 
For MIA, we applied the methods Vandp and Neighb, denoted as~\modelname-v and~\modelname-n when inserted into our method. 
For AIA, we introduced the PPGL method, similarly represented as~\modelname~p in~\modelname. 
It is evident that most~\modelname~variants disrupt existing defense mechanisms to varying extents. 
In particular, MIA achieves an increase in attack AUC-ROC of up to 18.7\% on the Cora dataset using GCN. Notably,~\modelname's AIA demonstrates exceptional attacking power on the sparsely distributed Pokec-n dataset, while Vanilla can only maintain random guess results.

\noindent \textbf{Case study.}
We aim to answer: Does~\modelname~successfully obfuscate itself so the target model is unaware during training? 
We utilized a radar chart, as illustrated in Fig.~\ref{fig:casestudy}, to comprehensively demonstrate the impact of~\modelname~on $\mathcal{F}_T$ during MIA. 
~\modelname~distorts the originally even distribution of the polygon into a conical shape, revealing the differences in training accuracy ($\mathcal{F}_T^{tr}$, $\mathcal{F}_S^{tr}$) and testing accuracy ($\mathcal{F}_T^{te}$, $\mathcal{S}_T^{te}$) of $\mathcal{F}_T$ and $\mathcal{F}_S$ under attack. This indicates that~\modelname~obfuscates the target's training process and increases the target's overfitting. 
Additionally, we use $\mathcal{F}_A^{(0)}$ and $\mathcal{F}_A^{(1)}$ to denote the true negative and true positive of $\mathcal{F}_A$'s binary classification. 
Notably, \modelname~decreases the misclassification rate.

\begin{figure}[t]
\centering
\includegraphics[width=1\columnwidth]{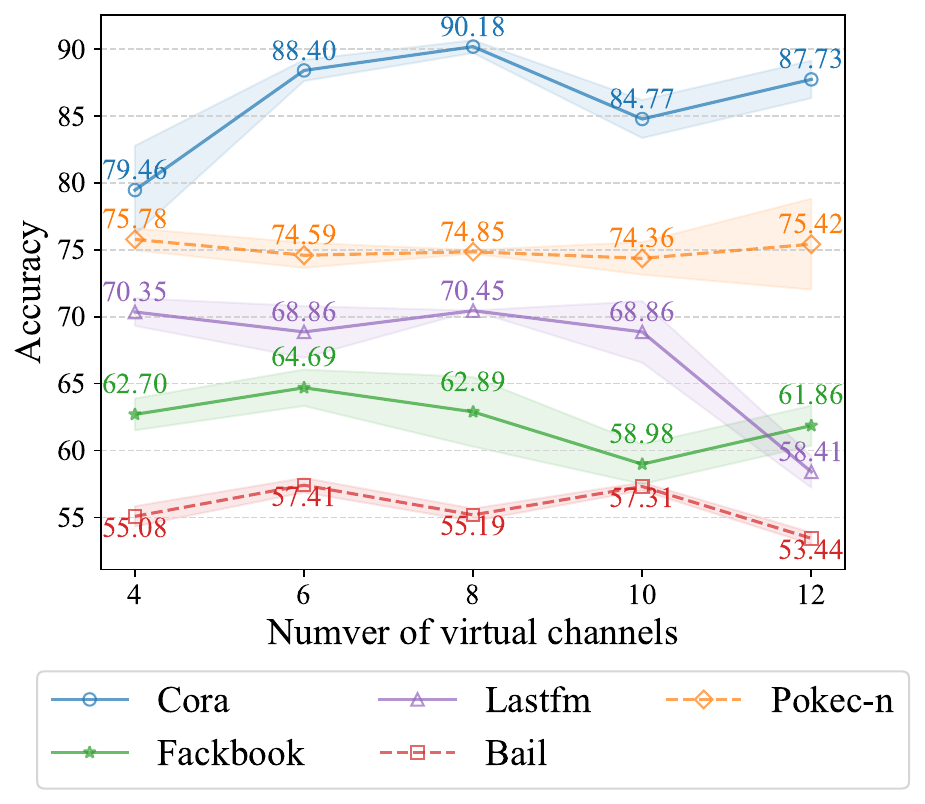}
\centering
\caption{Hyperparameter analysis.}
\label{fig:hyperpara}
\end{figure}

\noindent \textbf{Hyperparameter analysis.} 
We conduct the hyperparameter analysis for virtual channels $k$ in the disentanglement mechanism to verify the role of latent factors as key elements affecting the isolation of prompt feature structures. 
The results are shown in Fig.~\ref{fig:hyperpara}, indicating that the sensitivity to attack accuracy when setting $k$ varies across different datasets. The Cora has higher feature dimensions, and the Facebook has a more complex topology. During learning latent factors, the decision boundaries are affected by more interference sources, necessitating attention to the setting and adjustment of channel numbers. 
In contrast, the attack accuracy of other datasets is less affected. 
In summary, the hyperparameter settings can be further optimized according to the characteristics of different datasets to improve the attack model's overall performance.

\section{Conclusion}
In this paper, we present a novel framework named~\modelname~to explore the role of prompts in inference attacks and enhance the adaptiveness of such attacks under unified prompt guidance. \modelname~initially extracts critical topological information from graph data during pre-training while concealing the malicious intent of prompt features. 
Subsequently, it generates posterior supervised attack data by querying target and shadow models with prompts. 
Finally,~\modelname~improves the localization of prompt knowledge through a disentanglement mechanism in downstream tasks. 
Extensive experiments have demonstrated that \modelname~exhibits superior inference attack capabilities and adaptability. 
Future optimizations will focus on reducing the computational overhead in pre-training structures.

\clearpage

\section*{Acknowledgments}
The corresponding author is Chunming Hu. This paper is supported by the STI 2030-Major Projects under Grant No. 2022ZD0120203 and the National Natural Science Foundation of China through Grant No. U21A20474, No. 62462007, and No. 62302023. We owe sincere thanks to all authors for their valuable efforts and contributions. 
\bibliography{ref}
\appendix

\counterwithin{table}{section}
\counterwithin{figure}{section}
\counterwithin{equation}{section}

\clearpage
\section{Algorithm}\label{sec:Alg}

\begin{algorithm}
\caption{\modelname}
\label{alg:algorithm}
\textbf{Input}: Graph $G=(V,E,\mathbf{X})$; Non-linear rectifier $\tau $; Label of classification $\mathbf{Y} ^C$; Activation function $\mathbf{Y} ^C$.\\
\textbf{Parameter}: Number of layer $L$; Hyperparameters $t,k,\beta_A.\beta_M, \alpha,t$\\
\textbf{Output}: Predicted label $\mathcal{Y} ^{te}$;
\begin{algorithmic}[1] 

\STATE Initialize model parameters $\mathbf{W}, \mathbf{w}, \Theta, \theta $; 
\STATE Sample $G_{lo}(\hat{x} )$ using $t$ and shuffled $x$; 
\STATE $\mathbf{Z}_{\mathbf{x} }^0\gets \mathbf{X}$;
\STATE Construct set $\mathcal{N}_{v,k} \gets \left \{ u\in V|d(u,v)=k \right \} $;
\WHILE{\textit{not converge}}
\FOR {$l=1,\dots ,L$ and $v\in V$} 
\STATE $\hat{\mathbf{Z}}_{\mathbf{x} ,v}^{l-1} \leftarrow \tau (\mathbf{Z}_{\mathbf{x} ,v}^{l-1}) \mathbf{W}^{l}$;
\STATE $\phi_{v,k}^{l} \leftarrow \sigma (\{(\mathbf{\hat{Z} } _{\mathbf{x} , v}^{l-1} ||\hat{\mathbf{Z} }_{\mathbf{x} , u}^{l-1}) \mathbf{w} ^{T}\}_{u \in \mathcal{N}_{v,k}})$;
\STATE $\hat{\mathbf{Z}} _{At,Me}^l \gets \cup_{v \in \mathcal{N}_{v,k}}  \{ u   \in  \mathcal{N}_{v,k}  \mid   u  \sim  \operatorname{Gaussian}(\phi_{v, k}^{l})\}$;
\STATE $\mathbf{Z}^{l}_{\mathbf{x} ,v}\gets\sum_{(u,v)\in\hat{\mathbf{Z}} _{At,Me}^l}\hat{\mathbf{Z}}^{l-1}_{\mathbf{x} ,v}$
\ENDFOR
\STATE Get global representations $\mathbf{z}_{gl}\leftarrow \mathbf{Z}^{l}_{\mathbf{x} ,v}$ and $\mathbf{\hat{z} }_{gl}\leftarrow \mathbf{Z}^{l}_{\mathbf{\hat{x} } ,v}$ using Eq.~\eqref{eq:bilinear} and Eq.~($13$);
\STATE Compute the pre-training learning objective and update $\theta$ using Eq.~\eqref{eq:loss1};
\ENDWHILE
\STATE Train $\mathcal{F}_T$ and $\mathcal{F}_S$ with $\mathbf{p}$ using Eq.~\eqref{eq:prompt};
\STATE Generate $G_A, \mathbf{h}$ with $\mathcal{F}_T$ and $\mathcal{F}_S$. 
\WHILE{\textit{not converge}}
\FOR{$t=1,\dots ,T$}
\STATE $\mathbf{z}_k\gets$ Mapping $\mathbf{p}$ into channel $t$ using Eq.~\eqref{eq:disenmapping}; 
\STATE  $\hat{\mathbf{z}}_k\gets$Re-weighting the edge probability using Eq.~\eqref{eq:disen_reweigt};
\STATE  Compute node's disentangled representations $\mathbf{d}_k$ using Eq.~\eqref{eq:disen_repres};
\ENDFOR
\STATE Guiding attack inference using Eq.~\eqref{eq:loss2};
\STATE Update $\Theta$.
\ENDWHILE
\end{algorithmic}
\end{algorithm}

\section{Proofs} \label{sec:proof}
\subsection{Proof of Proposition 1}
We restate Proposition~\ref{pr:prop1}: 
\begin{Propapx} \label{pr:prop1apx}
(The lower bound of $I(Z_X; Y^C)$).
For any variational distributions $ \mathbb{Q}_{1}\left(\mathbf{Y} ^C \mid \mathbf{Z} _{\mathbf{X} }^{L}\right)$ and $\mathbb{Q}_{2}(\mathbf{Y} ^C)$: 
\begin{equation}
\begin{aligned}
I\left(\mathbf{Y} ^C ; \mathbf{Z} _{\mathbf{X} }^{L}\right) \geq 
1
+\mathbb{E}\left[\log \frac{ \mathbb{Q}_{1}\left(\mathbf{Y} ^C \mid \mathbf{Z} _{\mathbf{X} }^{L}\right)}{\mathbb{Q}_{2}(\mathbf{Y} ^C)}\right]& \\
-\mathbb{E}_{\mathbb{P}(\mathbf{Y} ^C) \mathbb{P}\left(\mathbf{Z} _{\mathbf{X} }^{L}\right)}\left[\frac{  \mathbb{Q}_{1}\left(\mathbf{Y} ^C \mid \mathbf{Z} _{\mathbf{X}}^{L}\right)}{\mathbb{Q}_{2}(\mathbf{Y} ^C)}\right]&
\end{aligned}
\end{equation} 
\end{Propapx} 

We employ the variational bounds of mutual information $I_{NWJ}$ as proposed by ~\citet{nguyen2010estimating}, with comprehensive details provided in~\citet{poole2019variational, wu2020graph}. 

\begin{Lemma}   
For any two random variables $\mathbf{Y},\mathbf{Z}$ and any function $f(\mathbf{Y},\mathbf{Z})$, we have:

\begin{equation}
    \begin{aligned}
        I(\mathbf{Y},\mathbf{Z}) \ge & \mathbb{E}_{\mathbb{P}(\mathbf{Y},\mathbf{Z})} \left [ f(\mathbf{Y},\mathbf{Z}) \right] \\
        &- \mathrm{e}^{-1}\mathbb{E}_{\mathbb{P}(\mathbf{Y})\mathbb{P}(\mathbf{Z})} \left[ \mathrm{e}^{f(\mathbf{Y},\mathbf{Z})} \right].
    \end{aligned}
\end{equation}
\end{Lemma}
We use the above lemma to $\left(\mathbf{Y}^C , \mathbf{Z} _\mathbf{X} ^{(L)}\right)$ and plug with:
\begin{equation}
    \left(\mathbf{Y}^C , \mathbf{Z} _\mathbf{X} ^{(L)}\right)=1+\log \frac{\prod_{v \in V} \mathbb{Q}_1\left(\mathbf{Y} _v^C \mid \mathbf{Z} _{\mathbf{X} , v}^{(L)}\right)}{\mathbb{Q}_2(\mathbf{Y}^C )}.
\end{equation}
We conclude the proof of Proposition 1.

\subsection{Proof of Proposition 2}

\begin{Propapx}\label{pr:prop2apx} 
(The upper bound of $I(G;\mathbf{Z}_{At},\mathbf{Z}_{Me})$).
Two randomly selected indices $\mathcal{I}_ {a}, \mathcal{I}_ {m} \subset L$, are independent ($G \perp \mathbf{Z}_{At}^{L} \mid \left\{\mathbf{Z}_{At}^{l_a}\right\} \cup\left\{\mathbf{Z}_{Me}^{l_m}\right\}$) based on the aforementioned Markov chain, where $l_a \in \mathcal{I} _a$, $l_m \in \mathcal{I} _m$. 
Then for any variational distributions $\mathbb{Q}\left(\mathbf{Z}_{At}^{l}\right)$ and $\mathbb{Q}\left(\mathbf{Z}_{Me}^{l}\right)$:

\begin{equation}
\begin{aligned}
I(G;\mathbf{Z}_{At},\mathbf{Z}_{Me})
&\leq I \left(G;\left\{\mathbf{Z}_{At}^{l_a}\right\} \cup\left\{\mathbf{Z}_{Me}^{l_m}\right\}\right) \\
&\leq \sum_{l \in \mathcal{I}_a} \mathcal{T} _{At}^{l}+\sum_{l \in \mathcal{I}_m} \mathcal{T} _{Me}^{l} ,
\end{aligned}
\end{equation} 
\begin{align}
\text {where } 
&
\mathcal{T} ^{l}_{At}=\mathbb{E}\left[\log \frac{\mathbb{P}\left(\mathbf{Z}_{At}^{l} \mid \mathbf{Z}_{At}^{l-1}, \mathbf{Z}_{Me}^{l}\right)}{\mathbb{Q}\left(\mathbf{Z}_{At}^{l}\right)}\right]
, \\
&
\mathcal{T} ^{l}_{Me}=\mathbb{E}\left[\log \frac{\mathbb{P}\left(\mathbf{Z} _{Me}^{l} \mid \mathbf{A}, \mathbf{Z} _{At}^{l-1}\right)}{\mathbb{Q}\left(\mathbf{Z}_{Me}^{l}\right)}\right]
.
\end{align}

\end{Propapx} 
We apply the Data Processing Inequality (DPI)~\cite{beaudry2011intuitive} and the Markovian dependency to prove the first inequality. 

\begin{Lemma}
Given any three variables $\mathbf{X}$, $\mathbf{Y}$ and $\mathbf{Z}$, which follow the Markov Chain $<\mathbf{X} \to \mathbf{Y} \to \mathbf{Z}>$, we have:
\begin{equation}
    I(\mathbf{X};\mathbf{Y}) \ge I(\mathbf{X};\mathbf{Z}).
\end{equation}
\end{Lemma}
For the second inequality, 
we have Markovian sequences $\mathcal{S}_\mathbf{m}$ and $\mathcal{S}_\mathbf{a}$, and
we can see upper bounds proof according to~\citet{wu2020graph} for $I\left(G;\left\{\mathbf{Z}_{At}^{l}\right\}_{l \in \mathcal{I}_a} \right)$ and $I\left(G;\left\{\mathbf{Z}_{Me}^{l}\right\}_{l \in \mathcal{I}_m}\right))$.

\section{Experiments}
We provide the following additional description of the~\modelname~experiment
\subsection{Dataset} \label{sec:Dataset}
The statistics of the dataset are shown in Tab.~\ref{tab:dataset_description}, and the details of the dataset we used are as follows.
\textit{For the MIA}, we set three datasets.
\begin{itemize}
\item \textbf{Cora} is a citation network of academic papers in machine learning, where nodes represent papers and edges represent citation relationships between them.
\item \textbf{Facebook} describes the social network of relationships between social media pages collected from user connections. 
Nodes represent users, and edges represent friendships between them. 
Nodes may also represent brands, groups, celebrities, and other public pages, with edges indicating mutual follows or interactions. 
\item \textbf{Lastfm} is a social network reflecting users' musical interests. Nodes can represent users or music tags, and edges denote the interesting relationship between users and tags or the social relationships between users.
\end{itemize}
\textit{For the AIA}, we utilize two datasets that are labeled with sensitive attributes.
\begin{itemize}
\item \textbf{Bail} is a U.S. bail dataset recording the defendants' attribute information and bail decisions in cases. 
Nodes represent the defendants and edges denote associations between cases. The sensitive attribute is race, and node labels indicate whether the defendant is likely to commit a violent crime.
\item \textbf{Pokec-n} is a social network extracted from the Slovakian social network Pokec. Nodes represent users, edges are friendships between users, the sensitive attribute is the region, and node labels indicate the job type of the users.
\end{itemize}

\begin{table}
\centering

\begin{tabular}{cl|rrrr}
\toprule
\multicolumn{2}{c|}{\textbf{Dataset}}  & \textbf{\#Node} & \textbf{\#Edge} & \textbf{\#Feat}& \textbf{\#Class} \\ 
\midrule
&\textbf{Cora}       &2,708 &5,278 &1,433 & 7  \\
&\textbf{Facebook}   &22,470 &171,002 &128 & 4   \\
&\textbf{Lastfm}     &7,624  &27,806  &128  &18  \\
\midrule
&\textbf{Bail}       &18,876  &311,870  &18  &2  \\
&\textbf{Pokec-n}    &66,569  &517,047  &264  &5  \\
\bottomrule
\end{tabular}

\caption{Statistics of datasets.}
\label{tab:dataset_description}
\end{table}

\subsection{Baseline.} \label{sec:baseline}
A description of the backbone model used is as follows:
\begin{itemize}
\item \textbf{GCN} is one of the first deep learning models applied to graph-structured data that propagates node features across the graph by extending the concept of convolutional neural networks. 
\item\textbf{GAT} introduces an attention mechanism into the feature aggregation process, enabling the self-learning and adjustment of weights through backpropagation. 
\item\textbf{SAGE} denotes GraphSAGE, a sampling and aggregation-based graph embedding model that can train on graphs in batch mode. 
\item \textbf{STABLE}, an additional model, demonstrates the utility of~\modelname~in extracting hint information from robust models and is based on a contrastive learning framework.
Furthermore, we validated~\modelname's capability in defense models. 
\item \textbf{Vandp} aims to add Gaussian noise to the output of $\mathcal{F}_T$. 
\item \textbf{Neigbh} introduces query neighborhood perturbations to $\mathcal{F}_T$. 
\item \textbf{PPGL} separates non-sensitive attribute representations for downstream inference by learning latent representations of attributes.
\end{itemize}

\subsection{Settings} \label{sec:setting}
The evaluation metrics we have chosen are: 
$(1)$ \textbf{accuracy} measures the overall performance of the attack by providing a simple and intuitive metric. 
$(2)$ \textbf{F1 score} balances precision and recall, particularly useful in imbalanced data contexts. 
$(3)$ \textbf{AUC-ROC} assesses the model's predictive ability across various thresholds. 
In addition, hardware and software configurations for our experiments:
\begin{itemize}
    \item Ubuntu 20.04.6 LTS operating system.
    \item Intel(R) Xeon(R) Gold 6330 CPU @ 2.00GHz processor, 128G memory and A800 80G GPU.
    \item Software: CUDA 11.8.0, Python 3.9.20, PyTorch 2.4.0, PyTorch-Geometrics 2.6.1.
\end{itemize}

\begin{figure}[t]
\centering
\includegraphics[width=1\columnwidth]{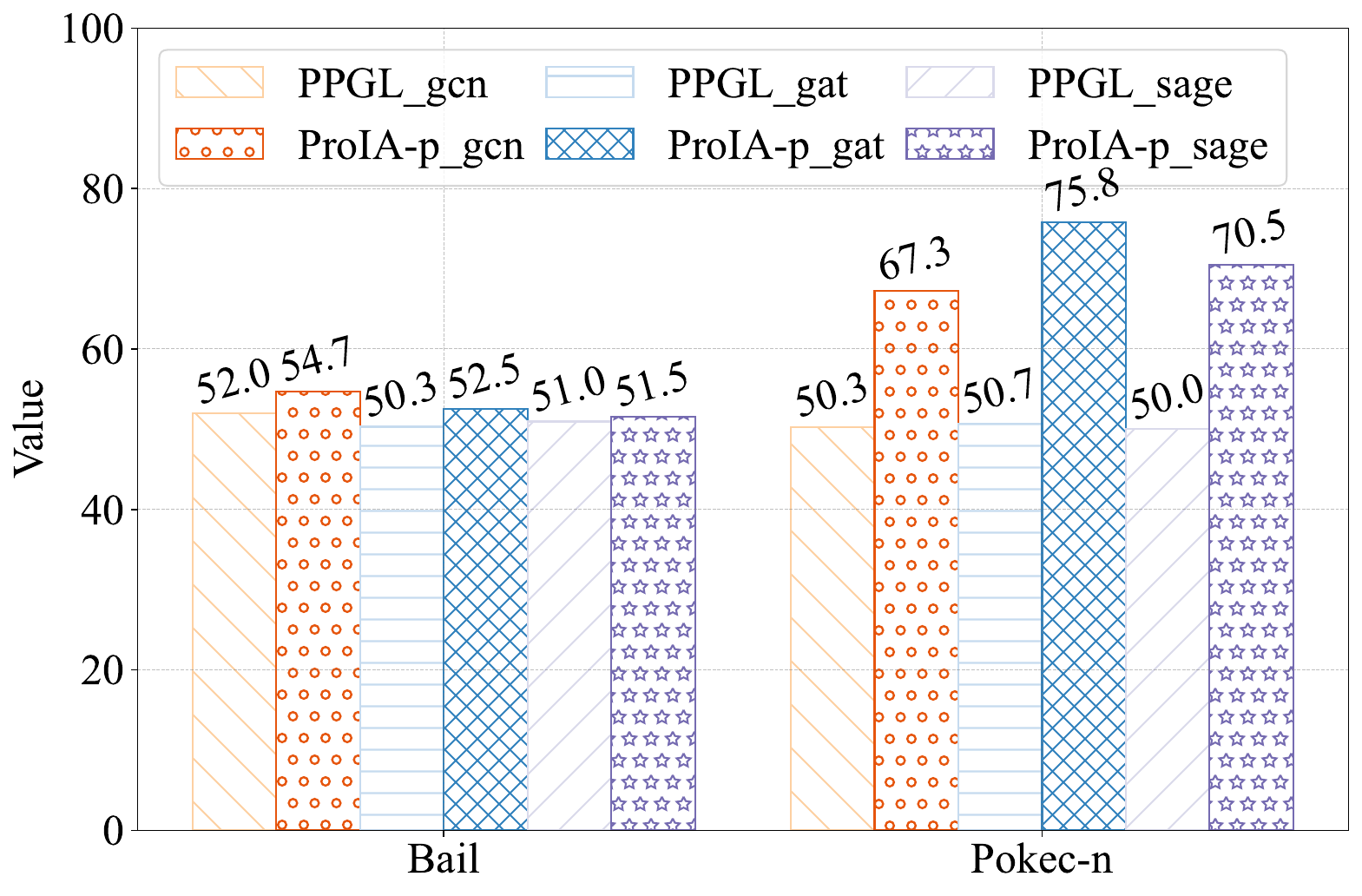}
\centering
\caption{Attack AUC-ROC scores on PPGL for AIA.}
\label{fig:defend_AIA}
\end{figure}

\begin{figure}[t]
{
\subfigure[Bail.]{
    \begin{minipage}{0.22\textwidth}
        \centering      
        \includegraphics[width=1\textwidth]{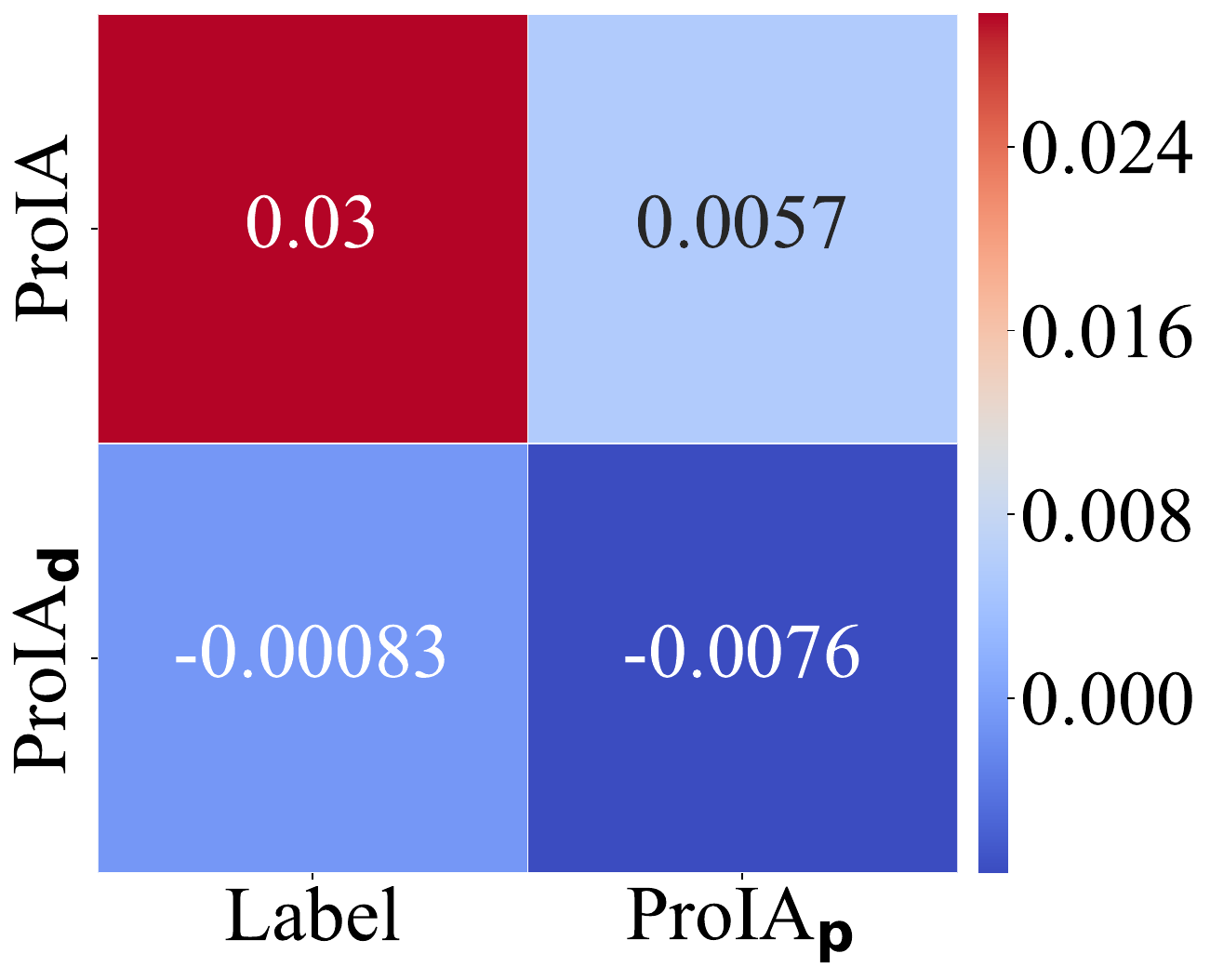}
        \label{fig:cc}
    \end{minipage}
}
\subfigure[Pokec-n.]{
    \begin{minipage}{0.22\textwidth}
        \centering
        \includegraphics[width=1\textwidth]{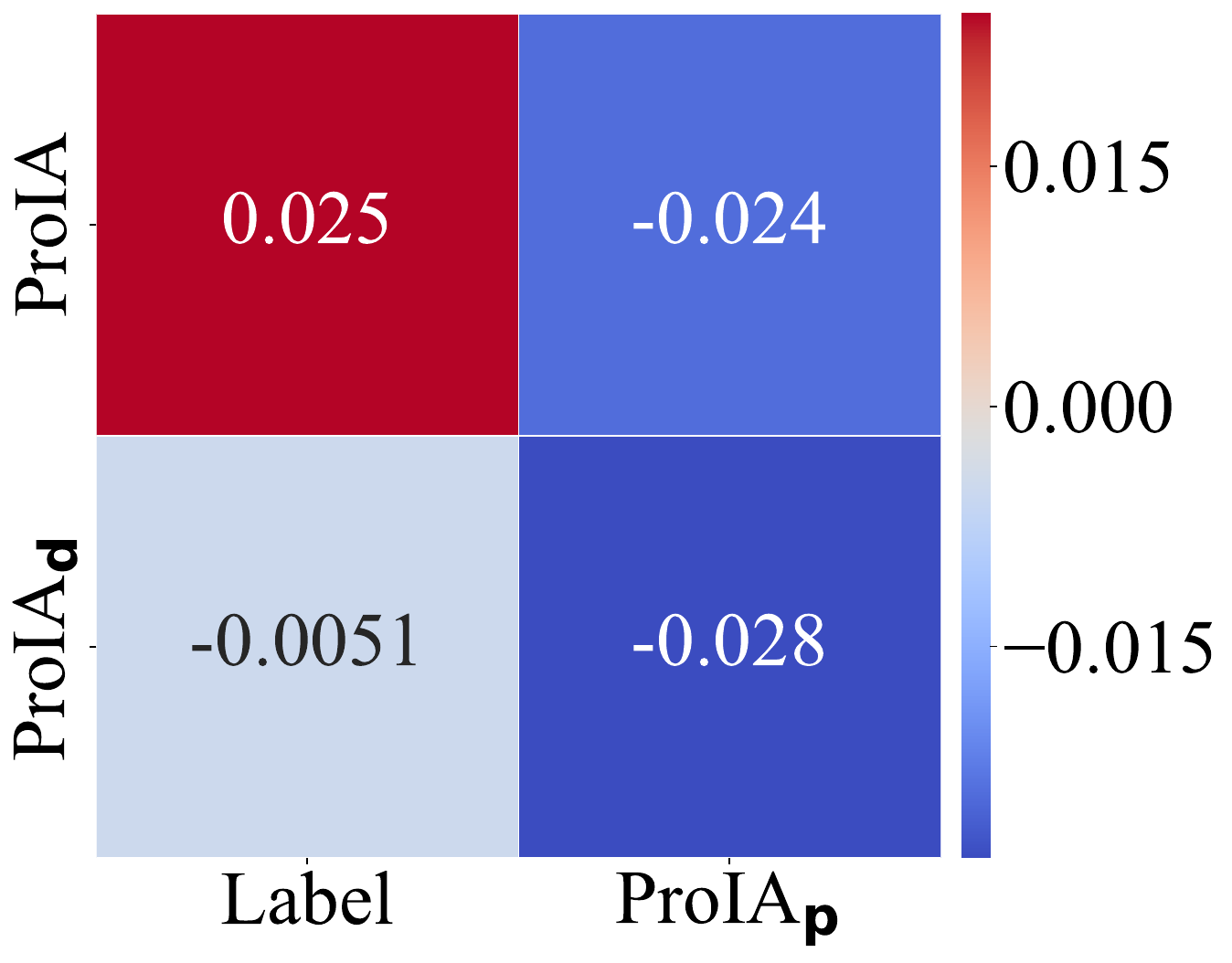}
        \label{fig:cd}
    \end{minipage}
}

\centering
\caption{Ablation study for the disentanglement mechanism on AIA. The colors from dark to light represent the degree of correlation. }
\label{fig:ablationstudyAIA}
}
\end{figure}

\subsection{Additional Results}
Additional results are presented in Fig.~\ref{fig:defend_AIA} and analyzed in section~\ref{sec:performance}. 
Fig.~\ref{fig:ablationstudyAIA} further illustrates the relationship between the disentanglement mechanism and the attack task. 
We record the attack posteriors (\modelname~and~\modelname$_\mathbf{d}$) and prompt features (\modelname$_\mathbf{p}$) of the attack test set, then calculate their correlation with the labels after performing Principal Component Analysis. 
It can be observed that~\modelname~exhibits high adaptability, and~\modelname$_\mathbf{d}$ maintains some effectiveness even when disentanglement.

\end{document}